\documentclass[11pt,a4paper]{article}
\usepackage[hyperref]{naaclhlt2018}

\usepackage{times}
\usepackage{latexsym}

\usepackage{url}

\usepackage{amsmath}
\usepackage{multirow}
\usepackage{url}

\aclfinalcopy 
\def\aclpaperid{593} 

\usepackage{xspace}
\usepackage{array}
\usepackage{graphicx}
\usepackage{booktabs}
\usepackage{multirow}
\usepackage{booktabs}
\usepackage{tabularx}
\usepackage{qtree}
\usepackage{graphicx}
\usepackage{tabularx}
\usepackage{bm}
\usepackage{xspace}
\usepackage{paralist}
\usepackage{comment}
\usepackage{enumitem}
\usepackage{url}
\usepackage{color}

\newcommand{\featSet}[1]{{\textsc{#1}}}

\newcommand{\featSetNgrams}[0]{\featSet{Lexical}\xspace}

\newcommand{\featSetNgramsShort}[0]{\featSet{LEX}\xspace}

\newcommand{\featSetMetaDataShort}[0]{\featSet{THR}\xspace}

\newcolumntype{M}{>{\centering\let\newline\\\arraybackslash\hspace{0pt}}X}
\newcolumntype{Y}{>{\centering\arraybackslash}X}

\newcommand{\CBCount}{\textit{CBCount}\xspace}
\newcommand{\NCBCount}{\textit{NCBCount}\xspace}
\newcommand{\NACount}{\textit{NACount}\xspace}
\newcommand{\ROBCount}{\textit{ROBCount}\xspace}

\newcommand{\cbcb}{\textsc{CommittedBelief}\xspace}
\newcommand{\cbncb}{\textsc{NonCommittedBelief}\xspace}
\newcommand{\cbna}{\textsc{NonApplicable}\xspace}
\newcommand{\cbrob}{\textsc{ReportedBelief}\xspace}

\newcommand{\cbcbshort}{\textsc{CB}\xspace}
\newcommand{\cbncbshort}{\textsc{NCB}\xspace}
\newcommand{\cbnashort}{\textsc{NA}\xspace}
\newcommand{\cbrobshort}{\textsc{ROB}\xspace}

\newcommand{\baseline}{\textit{BaseLine}\xspace}

\newcommand{\lnbaseline}{\textit{LN}\xspace}
\newcommand{\pnbaseline}{\textit{PN}\xspace}
\newcommand{\mnbaseline}{\textit{MN}\xspace}

\usepackage{lingmac}
\usepackage{amsthm}
\theoremstyle{definition}
\newtheorem{hypothesis}{H.}
\newcommand{\hyporef}[1]{{H.\ref{#1}\xspace}}

\newcommand{\lnappend}{\textit{LN\textsuperscript{CBApnd}\xspace}}
\newcommand{\pnappend}{\textit{PN\textsuperscript{CBApnd}\xspace}}
\newcommand{\mnappend}{\textit{MN\textsuperscript{CBApnd}\xspace}}

\newcommand{\powerpredictor}{{\textsc{PowerPredictor}}\xspace}

\newcommand{\vanish}[1]{}

\usepackage[english]{babel}
\usepackage{blindtext}

\title{Author Commitment and Social Power: \\ Automatic Belief Tagging to Infer the Social Context of Interactions}

\author{Vinodkumar Prabhakaran\\
	Stanford University \\
	{\tt \small vinod@cs.stanford.edu} \\\And
	Premkumar Ganeshkumar\\
	Agolo, Inc.\\
	{\tt \small prem@agolo.com} \And
	Owen Rambow\\
	Elemental Cognition, Inc.\\
	{\tt \small owenr@elementalcognition.com} \\}

\date{}

\begin{document}
	\maketitle
	\begin{abstract}
		Understanding how social power structures affect the way we interact with one another is of great interest to social scientists who want to answer fundamental questions about human behavior, as well as to computer scientists who want to build automatic methods to infer the social contexts of interactions. 
In this paper, we employ advancements in extra-propositional semantics extraction within NLP to study how author commitment reflects the social context of an interactions. Specifically, we investigate whether the level of commitment expressed by individuals in an organizational interaction reflects the hierarchical power structures they are part of.
We find that subordinates use significantly more instances of non-commitment than superiors. More importantly, we also find that subordinates attribute propositions to other agents more often than superiors do --- an aspect that has not been studied before. Finally, we show that enriching lexical features with commitment labels captures important distinctions in social meanings.

 	\end{abstract}

\section{Introduction}
\label{sec:cb_intro}

Social power is a difficult concept to define, but is often manifested in how we interact with one another. Understanding these manifestations is important not only to answer fundamental questions in social sciences about power and social interactions, but also to build computational models that can automatically infer social power structures from interactions.
The availability and access to large digital repositories of naturally occurring social interactions and the advancements in natural language processing techniques in recent years have enabled researchers to perform large scale studies on linguistic correlates of power, such as words and phrases \cite{BramsenEPA11,Gilbert_2012}, linguistic coordination \cite{Danescu2012},  agenda control \cite{Taylor2012}, and dialog structure \cite{prabhakaran-rambow:2014:P14-2}.

Another area of research that has recently garnered interest within the NLP community is the modeling of author commitment in text. 
Initial studies in this area were done in processing hedges, uncertainty and lack of commitment, specifically focused on scientific text \cite{mercer2004frequency,di2006using,farkas-EtAl:2010:CoNLL-ST}.
More recently, researchers have also looked into capturing author commitment in non-scientific text, e.g., levels of factuality in newswire \cite{sauri/pustejovsky:2009}, types of commitment of beliefs in a variety of genres including conversational text \cite{diab-EtAl:2009:LAW-III,prabhakaran-EtAl:2015:starsembelief}.
These approaches are motivated from an information extraction perspective, for instance in aiding tasks such as knowledge base population.\footnote{The BeSt track of the 2017 TAC-KBP evaluation  aimed at detecting the ``belief and sentiment of an entity toward another entity, relation, or event'' (\url{http://www.cs.columbia.edu/~rambow/best-eval-2017/}).}
However, it has not been studied whether such sophisticated author commitment analysis can go beyond what is expressed in language and reveal the underlying social contexts in which language is exchanged.

In this paper, we bring together these two lines of research; 
we study how power relations correlate with the levels of commitment authors express in interactions. 
We use the power analysis framework built by \citet{prabhakaran-rambow:2014:P14-2} to perform this study, and  measure author commitment using the committed belief tagging framework introduced by \cite{diab-EtAl:2009:LAW-III} that distinguishes different types of beliefs expressed in text.
Our contributions are two-fold --- statistical analysis of author commitment in relation with power, and enrichment of lexical features with commitment labels to aid in computational prediction of power relations.
In the first part, we find that author commitment is significantly correlated with the social power relations between their participants --- subordinates use more instances of non-commitment, a finding that is in line with sociolinguistics studies in this area. 
We also find that subordinates use significantly more reported beliefs (i.e., attributing beliefs to other agents) than superiors. 
This is a new finding; to our knowledge, there has not been any sociolinguistics studies investigating this aspect of interaction in relation with power.
In the second part, we
present novel ways of incorporating the author commitment information into lexical features that can capture important distinctions in word meanings conveyed through the belief contexts in which they occur; distinctions that are lost in a model that conflates all occurrences of
a word into one unit.

We first describe the related work in computational power analysis and computational modeling of cognitive states in Section~\ref{sec:cb_related}. In Section~\ref{sec:cb_framework}, we describe the power analysis framework we use. Section~\ref{sec:researchquestion} formally defines the research questions we are investigating, and describes how we obtain the belief information. In Section~\ref{sec:cb_analysis}, we present the statistical analysis of author commitment and power. Section~\ref{sec:cb_power_experiments} presents the utility of enriching lexical features with belief labels in the context of automatic power prediction.
Section~\ref{sec:conclusion} concludes the paper and summarizes the results.

\section{Related Work}
\label{sec:cb_related}

The notion of belief that we use in this paper 
\cite{diab-EtAl:2009:LAW-III,prabhakaran-EtAl:2015:starsembelief} 
is closely related to the notion of factuality that is captured in FactBank \cite{sauri/pustejovsky:2009}. They capture three levels of factuality, certain (CT), probable (PB), and possible (PS), as well as the underspecified factuality (Uu).  They also record the corresponding polarity values,
and the source of the factuality assertions to distinguish between factuality assertions by the author and those by the agents/sources introduced by the author.
While FactBank offers a finer granularity, they are annotated on newswire text.
Hence, 
we use the corpus of belief annotations \cite{prabhakaran-EtAl:2015:starsembelief} that is obtained on online discussion forums, which is closer to our genre.

Automatic hedge/uncertainty detection is a very closely related task to belief detection.
The belief tagging framework we use aims to capture the cognitive states of authors, whereas hedges are linguistic expressions that convey one of those cognitive states --- non-committed beliefs. 
Automatic hedge/uncertainty detection has generated active research in recent years within the NLP community.
Early work in this area focused on detecting speculative language in scientific text \cite{mercer2004frequency,di2006using,kilicoglu2008recognizing}.
The open evaluation as part of the CoNLL shared task in 2010 to detect uncertainty and hedging in biomedical and Wikipedia text \cite{farkas-EtAl:2010:CoNLL-ST} triggered further research on this problem in the general domain \cite{agarwal2010detecting,morante2010memory,velldal2012speculation,choi2012hedge}.
Most of this work was aimed at formal scientific text in English. 
More recent work has tried to extend this work to other genres \cite{wei-EtAl:2013:Short,sanchezIWCS2015workshop} and languages \cite{velupillai2012shades,vincze:2014:Coling}, as well as building general purpose hedge lexicons \cite{prokofieva_hedging}. In our work, we use the lexicons from \cite{prokofieva_hedging} to capture hedges in text.

Sociolinguists have long studied the association between level of commitment and social contexts \cite{lakoff1973language,o1980women,hyland1998hedging}. 
A majority of this work studies
gender differences in the use of hedges, triggered by the influential work by Robin Lakoff \cite{lakoff1973language}. She argued that women use linguistic strategies such as hedging and hesitations in order to adopt an unassertive communication style, which she terms ``women's language''.
While many studies have found evidence to support Lakoff's theory (e.g., \cite{crosby1977female,preisler1986linguistic,carli1990gender}), there have also been contradictory findings (e.g., \cite{o1980women}) that link the difference in the use of hedges to other social factors (e.g., power). 
\newcite{o1980women} argue that the use of hedges is linked more to the social positions rather than gender, suggesting to rename ``women's language'' to ``powerless language''.
In later work, 
\newcite{o1982linguistic} formalized the notion of powerless language,
which formed the basis of many sociolinguistics studies on social power and communication.
\newcite{o1982linguistic} analyzed courtroom interactions and identified hedges and hesitations as some of the linguistic markers of ``powerless'' speech. 
However, there has not been any computational work which has looked into how power relations relate to the level of commitment expressed in text. 
In this paper, we use computational power analysis to perform a large scale data-oriented study on how author commitment in text reveals the underlying power relations.

There is a large body of literature in the social sciences that studies power as a social construct (e.g., \cite{French_Raven,Dahl_Power,Emerson_power,pfeffer1981power,wartenberg1990forms}) and how it relates to the ways people use language in social situations (e.g., \cite{BSMR:1951,bales1970personality,o1982linguistic,van1989structures,bourdieu1991language,ng1993power,fairclough2001language,locher2004power}).
Recent years have seen growing interest in computationally analyzing and detecting power and influence from interactions. 
Early work in computational power analysis used social network analysis based approaches \cite{Diesner05explorationof,Shetty2005,Creamer:2009} or email traffic
patterns \cite{Namata:2006}.
Using NLP to deduce social relations from online
communication is a relatively new area 
of active research.

\citet{BramsenEPA11} 
and \citet{Gilbert_2012} first applied NLP based techniques to predict power relations in Enron emails, approaching this task as a text classification problem using bag of words or ngram features.
More recently, our work has used dialog structure features derived from deeper dialog act analysis for the task of power prediction in Enron emails \cite{prabhakaran-rambow:2014:P14-2,prabhakaranCOLING2012-long,prabhakaranIJCNLP2013}.
In this paper, We use the framework of \cite{prabhakaran-rambow:2014:P14-2}, but we analyze a novel aspect of interaction that has not been studied before --- what level of commitment do the authors express in language.

There has also been work on analyzing power in other genres of interactions.
\citet{Strzalkowski:2010} and \citet{Taylor2012}  concentrate on lower-level constructs called {\em Language Uses} such as agenda control to predict power in Wikipedia talk pages.
\citet{Danescu2012} study how social power and linguistic coordination are correlated in Wikipedia interactions as well as Supreme Court hearings.
\citet{BracewellTW12} and \citet{SwayamdiptaAndRambow2012} try to identify pursuit of power in discussion forums.
\citet{biran-EtAl:2012:LSM} and \citet{rosenthal2014detecting} study the problem of predicting influence in Wikipedia talk pages, blogs, and other online forums.
\citet{prabhakaran-john-seligmann:2013:IJCNLP} study manifestations of power of confidence in presidential debates.

\section{Power in Workplace Email: Data and Analysis Framework}
\label{sec:cb_framework}

The focus of our study is to investigate whether 
the level of commitment participants express in their contributions in an interaction
is related to the power relations they have with other participants, and how 
it
can help in the problem of predicting social power.
In this section, we introduce the power analysis framework as well as the data we use in this study.

\subsection{Problem}

In order to model manifestations of power relations in interactions, we use our interaction analysis framework from \cite{prabhakaran-rambow:2014:P14-2}, where we introduced the problem of predicting organizational power relations between pairs of participants based on single email threads. 
The problem is formally defined as follows:
given an email thread $ \mathit{t} $, and a related interacting participant pair $ \mathit{(p_1, p_2)} $ in the thread, predict whether $ \mathit{p_1} $ is the \textit{superior} or \textit{subordinate} of $ \mathit{p_2} $.
In this formulation, a \textit{related interacting participant pair (RIPP)} is a pair of participants of the thread such that there is at least one message exchanged within the thread between them (in either direction) and that they are hierarchically related with a superior/subordinate relation.

\subsection{Data}

We use the same dataset we used in \cite{prabhakaran-rambow:2014:P14-2}, which is a version of the Enron email corpus in which the thread structure of email messages is reconstructed \cite{Yeh06emailthread}, and 
enriched by \newcite{apoorvOrgHP}
with gold organizational power relations, manually determined using
information from Enron organizational charts.  
The corpus captures dominance relations between 13,724 pairs of
Enron employees.  As in \cite{prabhakaran-rambow:2014:P14-2}, we use these dominance relation tuples to obtain gold labels for the \textit{superior} or \textit{subordinate} relationships between pairs of participants.
We use the same train-test-dev split as in \cite{prabhakaran-rambow:2014:P14-2}.
We summarize the number of threads and related interacting participant pairs in each subset of the data in Table~\ref{table:power_data}.

\begin{table}[h]
	\begin{center}
		\begin{tabular}{l  r r r}
\toprule
			\bf Description & \bf Train & \bf Dev  & \bf Test\\ 
\midrule
			Email threads  &  18079 & 8973 & 9144  \\
			\# of RIPPs & 7510 & 3578 & 3920  \\
\bottomrule
		\end{tabular}
		\caption{ \label{table:power_data} Data Statistics.
		Row 1: number of threads in subsets of the corpus. 
		Row 2: number of related interacting participant pairs
		in those subsets.
		RIPP: Related interacting participant pairs}
	\end{center}
\end{table}

\section{Research Hypotheses}
\label{sec:researchquestion}

Our first objective in this paper is to perform a large scale computational analysis of author commitment and power relations. Specifically, we want to investigate whether the commitment authors express towards their contributions in organizational interactions is correlated with the power relations they have with other participants. 
Sociolinguistics studies have found some evidence to suggest that lack of commitment expressed through hedges and hesitations is associated with lower power status \cite{o1982linguistic}.
However, in our study, we go beyond hedge word lists, and analyze different cognitive belief states expressed by authors using a belief tagging framework that takes into account the syntactic contexts within which propositions are expressed.

\subsection{Obtaining Belief Labels}

We use the committed belief analysis framework introduced by \cite{diab-EtAl:2009:LAW-III,prabhakaran-EtAl:2015:starsembelief} to model different levels of beliefs expressed in text. 
Specifically, in this paper, we use the 
4-way belief distinction --- \cbcb, \cbncb, \cbrob, and \cbna --- introduced in \cite{prabhakaran-EtAl:2015:starsembelief}.\footnote{We also performed analysis and experiments using an earlier 3-way belief distinction proposed by \cite{diab-EtAl:2009:LAW-III}, which also yielded similar findings.
We do not report the details of those analyses in this paper.}
\cite{prabhakaran-EtAl:2015:starsembelief} presented a corpus of online discussion forums with over 850K words, annotating each propositional head in text with one of the four belief labels. The paper also presented an automatic belief tagger 
trained on this data, which we use to obtain belief labels in our data.
We describe each belief label and our associated hypotheses below.

\noindent \paragraph{Committed belief (CB):} the writer strongly believes that the proposition is true, and wants the reader/hearer to believe that.
E.g.:
\eessmall{\label{deft_exa:cb}
	\item{John will {\bf submit} the report.}
	\item{I know that John is {\bf capable}.}
}

\noindent
As discussed earlier, lack of commitment in one's writing/speech is identified as markers of powerless language.  We thus hypothesize:

\begin{hypothesis} 
	\label{hypo_cb} \it
	Superiors use more instances of committed belief in their messages than subordinates.
\end{hypothesis}

\noindent \paragraph{Non-committed belief (NCB):} the writer explicitly identifies the
proposition as something which he or she could believe, but he or she
happens not to have a strong belief in,
for
example by using an epistemic modal auxiliary.  
E.g.:
\eessmall{\label{deft_exa:ncb}
	\item{John may {\bf submit} the report.}
	\item{I guess John is {\bf capable}.}
}

\noindent 
This class captures a more semantic notion of non-commitment than hedges,
since the belief annotation attempts to model the underlying meaning rather than language uses, and hence captures other linguistic means of expressing non-committedness.
Following \cite{o1982linguistic}, we formulate the below hypothesis:

\begin{hypothesis} 
	\label{hypo_ncb} \it
	Subordinates use more instances of non committed belief in their messages than superiors.
\end{hypothesis}

\noindent \paragraph{Reported belief (ROB):} the writer attributes belief
(either committed or non-committed) to another person or group. 
E.g.:
\eessmall{\label{deft_exa:rob}
	\item{Sara says John will {\bf submit} the report.}
	\item{Sara thinks John may be {\bf capable}.}
}

\noindent 
Note that this label is only applied when the writer's own belief in the proposition is unclear. 
For instance, if the first example above was \textit{Sara knows John will submit the report on-time}, the writer is expressing commitment toward the proposition that John will submit the report and it will be labeled as committed belief rather than reported belief. 
Reported belief captures instances where the writer is in effect limiting his/her commitment towards what is stated by attributing the belief to someone else. 
So, in line with our hypotheses for non-committed beliefs, we formulate the following hypothesis:

\begin{hypothesis} 
	\label{hypo_rob} \it
	Subordinates use more instances of reported beliefs in their messages than superiors.
\end{hypothesis}

\noindent \paragraph{Non-belief propositions (NA):} -- the writer expresses some other cognitive attitude toward the proposition, such as desire or
intention (\ref{deft_exa:na}a), or expressly states that he/she has no belief about the
proposition (e.g., asking a question (\ref{deft_exa:na}b)). 
E.g.:
\eessmall{\label{deft_exa:na}
	\item{I need John to {\bf submit} the report.}
	\item{Will John be {\bf capable}?}
}

\noindent
As per the above definition, requests for information (i.e., questions) and requests for actions are cases where the author is not expressing a belief about the proposition, but rather expressing the desire that some action be done. In the study correlating power with dialog act tags \cite{prabhakaran-rambow:2014:P14-2}, we found that superiors issue significantly more requests than subordinates. Hence, we expect the superiors to have significantly more non belief expressions in their messages, and 
formulate the following hypothesis:

\begin{hypothesis} 
	\label{hypo_na} \it
	Superiors use more instances of non beliefs in their messages than subordinates.
\end{hypothesis}

\subsection{Testing Belief Tagger Bias}

NLP tools are imperfect and may produce errors, which poses a problem when
using any NLP tool
for sociolinguistic analysis. More than the magnitude of error, we believe that whether the error is correlated with the social variable of interest (i.e., power) is more important; e.g., is the belief-tagger more likely to find ROB false-positives in subordinates’ text?  To test whether this is the case, we
performed manual belief annotation on around 500 propositional heads in our
corpus. Logistic regression test revealed that the belief-tagger is
equally likely to make errors (both false-positives and false-negatives,
for all four belief-labels) in sentences written by subordinates as
superiors (the null hypothesis accepted at $p>0.05$ for all eight tests). 

\section{Statistical  Analysis}
\label{sec:cb_analysis}

Now that we have set up the analysis framework and research hypotheses,
we present the statistical analysis of 
how superiors and subordinates differ in their relative use of expressions of commitment.

\subsection{Features}

For each participant of each pair of related interacting participants in our corpus, we 
aggregate each of the four belief tags:

\begin{itemize}[leftmargin=0.5cm,noitemsep]
	\item \CBCount: number of propositional heads tagged as Committed Belief (CB)
	\item \NCBCount: number of propositional heads tagged as Non Committed Belief (NCB)
	\item \ROBCount: number of propositional heads tagged as Reported Belief (ROB)
	\item \NACount: number of propositional heads tagged as Non Belief (NA)
\end{itemize}

\subsection{Hypotheses Testing}

Our general hypothesis is that power relations do correlate with the level of commitment people express in their messages;
i.e., at least one of \hyporef{hypo_cb} - \hyporef{hypo_na} is true. 
In this analysis, each participant of the pair $ \mathit{(p_1, p_2)} $ is a data instance.
We exclude the instances for which a feature value is undefined.\footnote{These are instances corresponding to participants who did not send any messages in the thread 
(some of the pairs in the set of related interacting participant pairs only had one-way communication) or whose messages were empty (e.g., forwarding messages).}

In order to test whether superiors and subordinates use different types of beliefs, we used a linear regression based analysis. For each feature, we built a linear regression model predicting the feature value using power (i.e., superior vs. subordinate) as the independent variable. 
Since verbosity of a participant can be highly correlated with each of these feature values (we found it to be highly correlated with subordinates \cite{prabhakaran-rambow:2014:P14-2}), we added token count as a control variable to the linear regression.

\begin{figure}
	\small
	\centering
	\includegraphics[width=\linewidth]{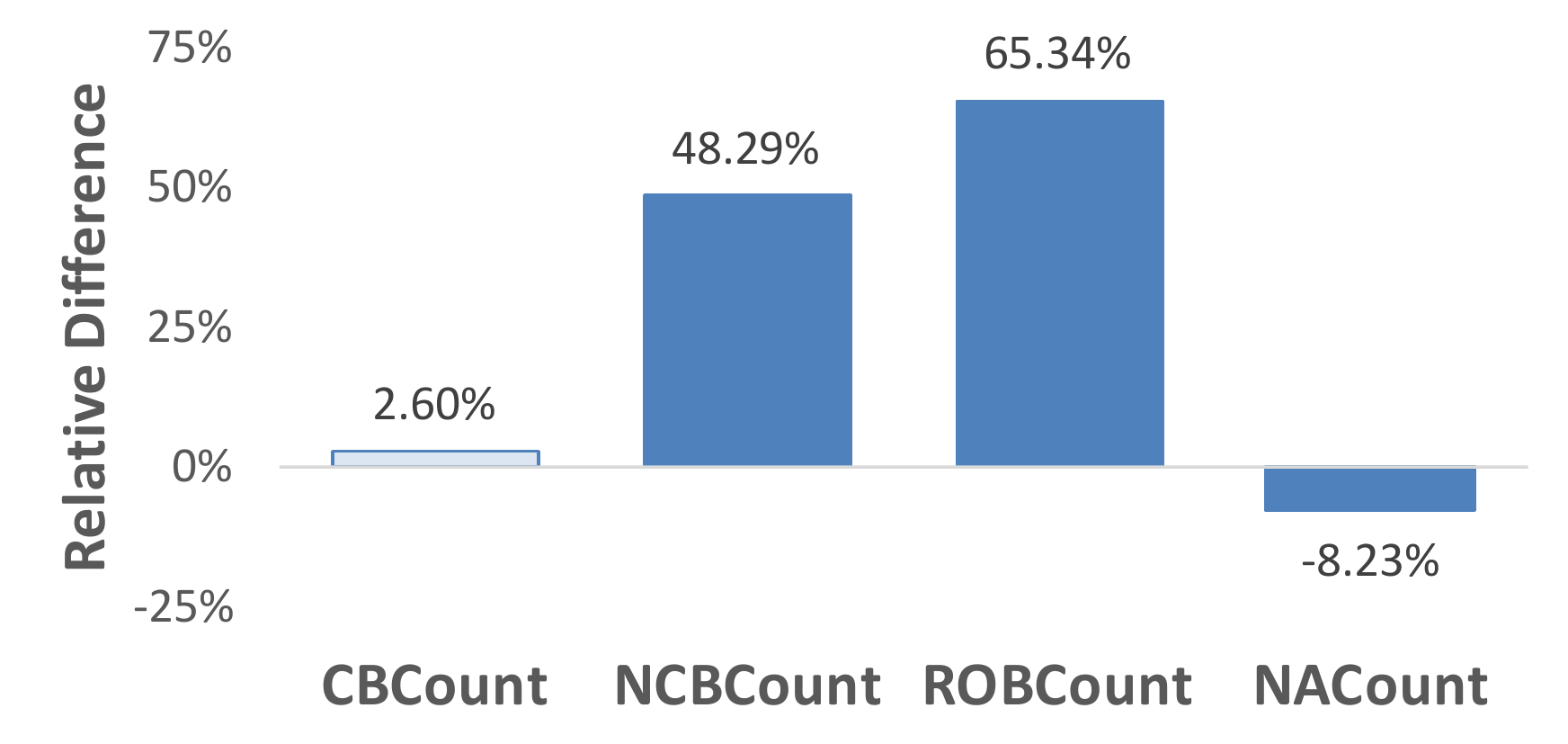}
	\caption{Relative difference (RD) between subordinates and superiors in their use of different types of commitment (counts normalized by word count of contributions). Dark bars: statistical significance at $ p<0.05 $.
	(RD = $\frac{(Mean(\textit{Subordinates}) - Mean(\textit{Superiors})) * 100} {Mean(\textit{Superiors})}$).
	\label{fig:relative_difference}}
\end{figure}

Our linear regression test revealed significant differences in NCB (b=-.095, t(-8.09), p$ < $.001), ROB (b=-.083, t(-7.162), p$ < $.001) and NA (b=.125, t(4.351), p$ < $.001), and no significant difference in CB (b=.007, t(0.227), p=0.821).
Figure~\ref{fig:relative_difference} pictorially demonstrates these results by plotting the difference between the mean values of each commitment feature (here normalized by token count) of superiors vs. subordinates, as a percentage of mean feature value of the corresponding commitment feature for superiors.
Dark bars denote statistically significant differences.

\subsection{Interpretation of Findings}

The results from our statistical analysis validate our original hypothesis that power relations do correlate with the level of commitment people express in their messages. This finding remains statistically significant ($ p<0.001 $) even after applying the Bonferroni correction for multiple testing.

The results on \cbncbshort confirm our hypothesis that subordinates use more non-committedness in their language. Subordinates' messages contain 48\% more instances of non-committed belief than superiors' messages, even after normalizing for the length of messages. This is in line with prior sociolinguistics literature suggesting that people with less power tend to use less commitment, previously measured in terms of hedges. However, in our work, we go beyond hedge dictionaries and use expressions of non-committedness that takes into account the syntactic configurations in which the words appear.

Another important finding is in terms of reported belief (\cbrobshort).
Our results strongly verify the hypothesis \hyporef{hypo_rob} that subordinates use significantly more reported beliefs than superiors. In fact, it obtained the largest magnitude of relative difference (65.3\% more) of all features we analyzed. 
To our knowledge, ours is the first study that analyzed the manifestation of power in authors attributing beliefs to others.
Our results are in line with the finding in \cite{Agarwal:2014:ECY:2639968.2640065} that ``if many more people get mentioned to a person then that person is the boss'', because as subordinates report other people's beliefs to superiors, they are also likely to mention them.

The finding that superiors use more NAs confirms our hypothesis \hyporef{hypo_na}. As discussed earlier, this is expected since superiors issue more requests (as found by \cite{prabhakaran-rambow:2014:P14-2}), the propositional heads of which would be tagged as \cbnashort by the belief tagger. However, our hypothesis \hyporef{hypo_cb} is proven false. Being a superior or subordinate does not affect how often their messages contain \cbcbshort, which suggests that power differences are manifested only in terms of lack of commitment.

\section{Commitment in Power Prediction}
\label{sec:cb_power_experiments}

Our next step is to explore whether we can utilize the hedge and belief labels to improve the performance of an automatic power prediction system. 
For this purpose, we use our \powerpredictor system \cite{prabhakaran-rambow:2014:P14-2} that predicts the direction of power between a pair of related interacting participants in an email thread. 
It uses a variety of linguistic and dialog structural features consisting of
verbosity features (message count, message ratio, token count, token ratio, and tokens per message), positional features (initiator, first message position, last message position), thread structure features (number of all recipients and those in the \textit{To} and \textit{CC} fields of the email, reply rate, binary features denoting the adding and removing of other participants), dialog act features (request for action, request for information, providing information, and conventional), and overt displays of power, and lexical features (lemma ngrams, part-of-speech ngrams, and
mixed ngrams, a version of
lemma ngrams with open class words replaced with their part-of-speech tags).
The feature sets are summarized in Table~\ref{table:cb_features} (\cite{prabhakaran-rambow:2014:P14-2} has a detailed description of these features).

\begin{table}[ht]
	\centering \small
	\begin{tabular}{c l }
		\toprule
		Set & Description \\
		\midrule
		VRB & Verbosity (e.g., message count) \\
		PST & Positional  (e.g., thread initiator?)\\
		THR & Thread structure (e.g., reply rate) \\
		DIA & Dialog act tagging (e.g., request count)\\
		ODP & Overt displays of power \\
		LEX & Lexical ngrams (lemma, POS, mixed ngrams) \\

		\bottomrule
	\end{tabular}
	\caption{\powerpredictor system: Features used\label{table:cb_features}}
\end{table}

None of the features used in \powerpredictor use information from the parse trees of sentences in the text
However, in order to accurately obtain the belief labels, deep dependency parse based features are critical \cite{Prabhakaran:2010}. 
We use the ClearTk wrapper for the Stanford CoreNLP pipeline to obtain the dependency parses of sentences in the email text.
To ensure an unified analysis framework, we also use the Stanford CoreNLP for tokenization, part-of-speech tagging, and lemmatization steps, instead of OpenNLP.
This change 
affects our analysis in two ways.
First, the source of part-of-speech tags and word lemmas is different from what was presented in the original system, which might affect the performance of the dialog act tagger and overt display of power tagger (DIA and ODP features). 
Second, we had to exclude 117 threads (0.3\%) from the corpus for which the Stanford CoreNLP failed to parse some sentences, resulting in the removal of 11 data points (0.2\%), only one of which was in the test set. 
On randomly checking, we found that  they contained non-parsable text such as dumps of large tables, system logs, or unedited dumps of large legal documents.

In order to better interpret how the commitment features help in power prediction, we use a linear kernel SVM in our experiments. Linear kernel SVMs are significantly faster than higher order SVMs, and our preliminary experiments revealed the performance gain by using a higher order SVM to be only marginal.  
We use the best performing feature set from \cite{prabhakaran-rambow:2014:P14-2} as a strong baseline for our experiments. This baseline feature set is the combination of thread structure features (THR) and lexical features (LEX). This baseline system obtained an accuracy of 68.8\% in the development set.

\subsection{Belief Label Enriched Lexical Features}

Adding the belief label counts into the SVM directly as features will not yield much performance improvements, as signal in the aggregate counts would be minimal given the effect sizes of differences we find in Section~\ref{sec:cb_analysis}. 
In this section, we investigate a more sophisticated way of incorporating the belief tags into the power prediction framework. Lexical features are very useful for the task of power prediction. However, it is often hard to capture deeper syntactic/semantic contexts of words and phrases using ngram features. We hypothesize that incorporating belief tags into the ngrams will enrich the representation and will help disambiguate different usages of same words/phrases. For example, let us consider two sentences: \textit{I need the report by tomorrow} vs. \textit{If I need the report, I will let you know}. The former is likely coming from a person who has power, whereas the latter does not give any such indication. Applying the belief tagger to these two sentences will result in 
\textit{I need(CB) the report ...} and \textit{If I need(NA) the report ...}.  
Capturing the difference between \textit{need(CB)} vs. \textit{need(NA)} will help the machine learning system to make the distinction between these two usages and in turn improve the power prediction performance.

In building the ngram features, whenever we encounter a token that is assigned a belief tag, 
we append 
the belief tag to the corresponding lemma or part-of-speech tag in the ngram. We call it the \textit{Append} version of corresponding ngram feature.
We summarize the different versions of each type of ngram features below:

\begin{itemize}[leftmargin=0.5cm,nosep]

\item \lnbaseline: the original word lemma ngram; e.g., \textit{i\_need\_the}.
\item \lnappend: word lemma ngram with appended belief tags; e.g., \textit{i\_need(CB)\_the}.
\item \pnbaseline: the original part-of-speech ngram; e.g., \textit{PRP\_VB\_DT}.
\item \pnappend: part-of-speech ngram with appended belief tags;  e.g., \textit{PRP\_VB(CB)\_DT}.
\item \mnbaseline: the original mixed ngram; e.g., \textit{i\_VB\_the}.
\item \mnappend: mixed ngram with appended belief tags;  e.g., \textit{i\_VB(CB)\_the}.

\end{itemize}

\begin{table}[t]
	\centering
	\begin{tabular}{l c}
		\toprule
		Feature Configuration in \featSetNgrams & Accuracy \\
		\midrule
		
		\lnbaseline+\pnbaseline+\mnbaseline (\baseline) & 68.8 \\
		\midrule
		\lnappend+\pnbaseline+\mnbaseline & \textbf{69.3 }\\

		\lnbaseline+\pnappend+\mnbaseline & 68.6 \\

		\lnbaseline+\pnbaseline+\mnappend & 69.0 \\

		\midrule

		\lnappend + \pnbaseline + \mnappend & 69.2 \\

		\bottomrule
	\end{tabular}
	\caption{\label{table:cb_power_results2}Power prediction results using different configurations of \featSetNgramsShort features. (The full feature set also includes \featSetMetaDataShort.)}
\end{table}

\noindent In Table~\ref{table:cb_power_results2},
we show the results obtained by incorporating the belief tags in this manner to the \featSetNgrams features of the original baseline feature set.
The first row indicates the baseline results and the following rows show the impact of incorporating belief tags 
using the \textit{Append} method. 
While the \textit{Append} version of both lemma ngrams and mixed ngrams improved the results, the \textit{Append} version of part of speech ngrams reduced the results. 
The combination of best performing version of each type of ngram obtained slightly lower result than using the \textit{Append} version of word ngram alone, which posted the overall best performance of 69.3\%, a significant improvement (p$<$0.05) over not using any belief information. 
We use the approximate randomization test \cite{Yeh:2000} for testing statistical significance of the improvement.

Finally, we verified that our best performing feature sets obtain similar improvements in the unseen test set. The baseline system obtained 70.2\% accuracy in the test set. The best performing configuration from Table~\ref{table:cb_power_results2} significantly improved this accuracy to 70.8\%. 
The second best performing configuration of using the \textit{Append} version of both word and mixed ngrams obtained only a small improvement upon the baseline in the test set.

\begin{figure*}[t]
\centering
\includegraphics[width=.98\linewidth]{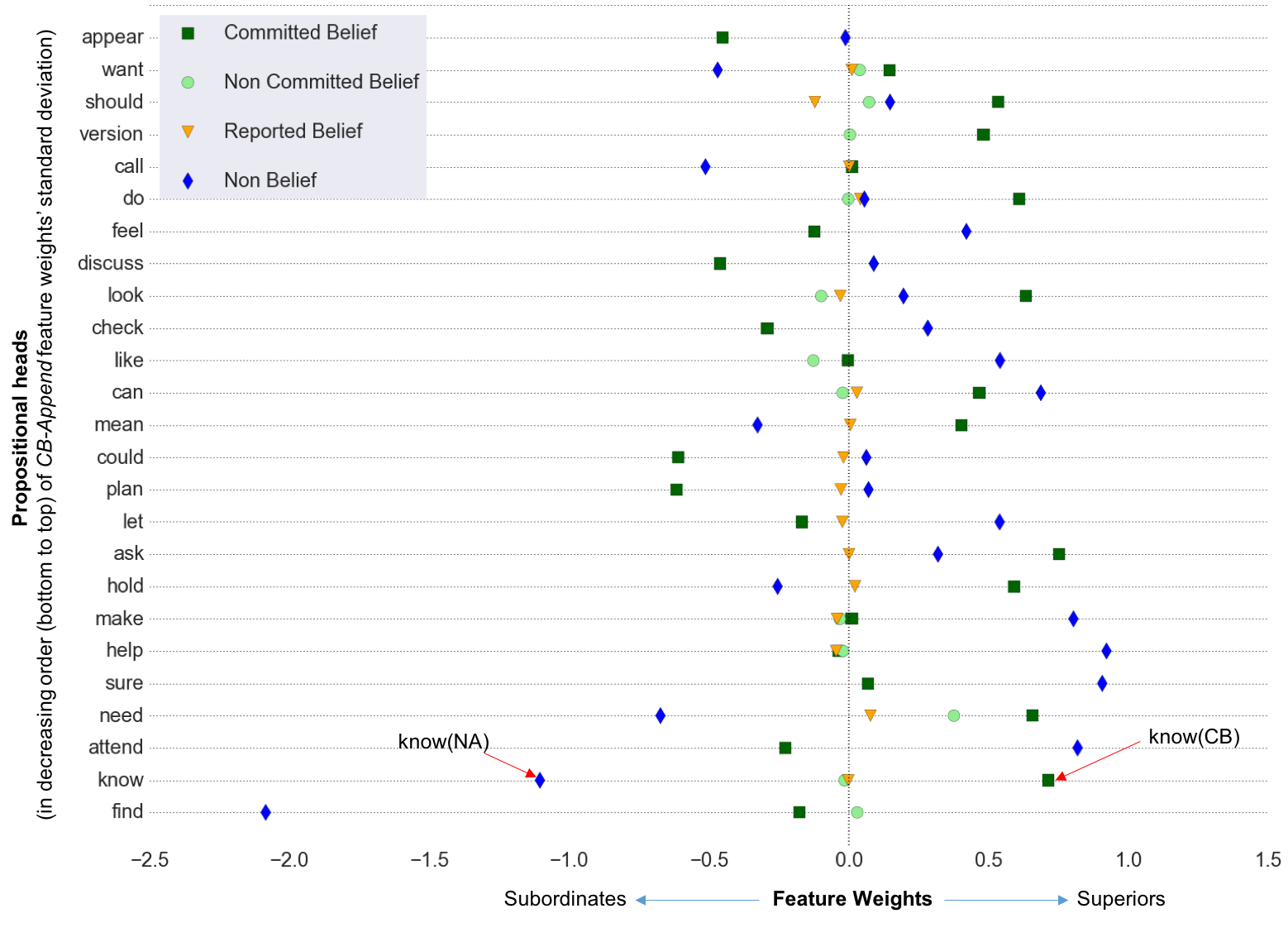}
\caption{\label{fig_cb_ngram_anal}Feature weights of different belief appended versions of 25 propositional heads whose lemma unigrams had the highest standard deviation. Y-axis denotes the propositional heads in decreasing order of standard deviation from bottom to top. X-axis denotes the feature weights. }
\label{fig:AllTags_25_LastRun}
\end{figure*}
 
\subsection{Word NGram Feature Analysis}
\label{sec:cb_ngram_anal}

We inspect the feature weights assigned to the \lnappend{} version of lemma ngrams in our best performing model. Each lemma ngram that contains a propositional head (e.g., \textit{need}) has four possible \lnappend{} ngram versions: \textit{need(CB)}, \textit{need(NCB)}, \textit{need(ROB)}, and \textit{need(NA)}. 
For each lemma ngram, we calculate the standard deviation of weights assigned to different \lnappend{} versions in the learned model as a measure of variation captured by incorporating belief tags into that ngram.\footnote{Not all lemma ngrams have all four versions; we calculated standard deviation using the versions present.} 

Figure~\ref{fig_cb_ngram_anal} shows the feature weights of different \lnappend{} versions of twenty five propositional heads whose lemma unigrams had the highest standard deviation. The y-axis lists propositional heads arranged in the decreasing order of standard deviation from bottom to top, while the x-axis denotes the feature weights. The markers distinguish the different \lnappend{} versions of each propositional head --- square denotes \cbcb, circle denotes \cbncb, triangle denotes \cbrob, and diamond denotes \cbna. The feature versions with negative weights are associated more with subordinates' messages, whereas those with positive weights are associated more with superiors' messages. 
Since \cbncbshort and \cbrobshort versions are rare, they rarely get high weights in the model.

We find that by incorporating belief labels into lexical features, we capture important distinctions in social meanings expressed through words that are lost in the regular lemma ngram formulation. 
For example, propositional heads such as \textit{know}, \textit{need}, \textit{hold}, \textit{mean} and \textit{want}
are indicators of power when they occur in \cbcbshort contexts (e.g., \textit{i need ...}), whereas their usages in \cbnashort contexts (e.g., \textit{do you need?}, \textit{if i need...}, etc.) are indicators of lack of power.
In contrast, the \cbcbshort version of \textit{attend}, \textit{let}, \textit{plan}, \textit{could}, \textit{check}, \textit{discuss}, and \textit{feel} (e.g., \textit{i will attend/check/plan ...}) are strongly associated with lack of power, while their \cbnashort versions (e.g., \textit{can you attend/check/plan?}) are indicators of power.

\section{Conclusion}
\label{sec:conclusion}

In this paper, we made two major contributions. First, we presented a large-scale data oriented analysis of how social power relations between participants of an interaction correlate with different types of author commitment in terms of their relative usage of hedges and different levels of beliefs --- committed belief, non-committed belief, reported belief, and non-belief.
We found evidence that subordinates use significantly more propositional hedges than superiors, and that superiors and subordinates use significantly different proportions of different types of beliefs in their messages. In particular, subordinates use significantly more non-committed beliefs than superiors. They also report others' beliefs more often than superiors.
Second, 
we investigated different ways of incorporating the belief tag
information into the machine learning system that automatically detects the
direction of power between pairs of participants in an interaction. 
We devised a sophisticated way of incorporating this information into the machine learning framework by appending the heads of propositions in lexical features with corresponding belief tags, demonstrating its utility in distinguishing social meanings expressed through the different belief contexts.

This study is based on emails from a single corporation, at the beginning of the 21st century.  Our findings on the correlation between author commitment and power may be reflective of the work culture that prevailed in that organization at the time when the emails were exchanged. It is important to replicate this study on emails from multiple organizations in order to assess whether these results generalize across board. It is likely that behavior patterns are affected by factors such as ethnic culture \cite{cox1991effects} of the organization, and the kinds of conversations interactants engage in (for instance, co-operative vs. competitive behavior \cite{hill1992cooperative}).
We intend to explore this line of inquiry in future work.

\section*{Acknowledgments}

This paper is partially based upon work supported by the DARPA DEFT program under a grant to Columbia University; all three co-authors were at Columbia University when portions of this work were performed. The views expressed here are those of the author(s) and do not reflect the official policy or position of the Department of Defense or the U.S. Government. 
We thank Dan Jurafsky and the anonymous reviewers for their helpful feedback.

\bibliographystyle{acl_natbib}
\bibliography{naaclhlt2018_final}

\end{document}